\documentclass[runningheads]{llncs}

\usepackage[T1]{fontenc}
\usepackage[utf8]{inputenc}
\usepackage[english]{babel}
\usepackage{graphicx}
\usepackage{amsmath}
\usepackage{amssymb}
\usepackage{algorithm}
\usepackage{algpseudocode}
\usepackage{booktabs}
\usepackage{multirow}
\usepackage{url}
\usepackage{hyperref}
\hypersetup{hidelinks}
\usepackage{placeins}

\begin{document}

\setlength{\textfloatsep}{5pt plus 1.0pt minus 1.0pt}

\setlength{\intextsep}{5pt plus 1.0pt minus 1.0pt}

\setlength{\floatsep}{5pt plus 1.0pt minus 1.0pt}

\title{Automated Tree Knowledge Graph Construction using Ontology Expansion and Retrieval from Vietnamese History Textbooks }


\author{Ket Doan Nguyen\inst{1}\orcidID{0009-0005-6698-9629} \and
Minh N. H. Nguyen\inst{1}}

\authorrunning{N. K. Doan and N. H. N. Minh}

\institute{The University of Danang, Vietnam-Korea University of Information and Communication Technology,
Danang, Vietnam\\
\email{\{nkdoan,nhnminh\}@vku.udn.vn}}

\titlerunning{Automated Tree KG Construction with Ontology Expansion}

\maketitle

\begin{abstract}
Hierarchical Knowledge graph (KG)-based retrieval augmented generation (RAG) has emerged as a powerful approach for supporting large language models with structured knowledge. However, there are primary challenges: (i)~the lack of methods for automatic KG construction using ontology expansion for low-resource languages such as Vietnamese, (ii)~the absence of systematic evaluation for knowledge retrieval strategies leveraging the hierarchical structures.
In this paper, we propose an end-to-end pipeline for \emph{KG construction} and retrieval strategies evaluation. 
In the \emph{KG construction}, we employ a three-phase hybrid relation extraction pipeline: intra-batch deduplication via Union-Find, approximate cross-batch search, and LLM extraction with a centroid filter that reduces prompts combined with a five-step dual-LLM validator to prevent bloated ontology.
A two-tier architecture consists of unmergeable structural nodes to preserve the document structure and mergeable content nodes.
The \emph{retrieval evaluation} consists of three graph traversal strategies: Top-Down, Horizontal, and Bottom-Up, which are evaluated on a synthetically generated benchmark of 1,210 Vietnamese queries from 109 subgraphs, categorized by five query directions.
In this paper, we construct the tree knowledge graph from Vietnamese high school History textbooks (nearly 400 pages) to produce 750 nodes and 4,341 semantic edges with controlled ontology growth from 40 to 41 types. Among experimental graph traversal strategies, the Top-Down strategy with structure surpasses the vector baseline by 4.7 percentage points in NDCG@10.
As a result, tree-structural information provides valuable information beyond flat cosine similarity but degrades performance when the query does not require structural context.

\end{abstract}

\section{Introduction}\label{sec:intro}

Vietnam's National Digital Transformation Program lists Education as one of eight priority sectors and calls for intelligent systems in teaching and learning. Among school subjects, History presents a distinct challenge. History lessons require students to trace connections across events, people, dates, and causal chains. Hence, reasoning over structured knowledge rather than memorizing isolated facts is mandatory for students to learn history more efficiently. Connected reasoning demands more than keyword matching against flat text and requires a knowledge representation for relations between entities. In Vietnamese language processing, there are existing emerging  pretrained models such as PhoBERT~\cite{Nguyen2020PhoBERT} that can perform well on multiple language tasks. To the best of our knowledge, no existing framework has been developed for constructing knowledge graphs from Vietnamese educational content, introducing an important gap for future research.

Knowledge graphs (KGs) store structured knowledge for supporting multi-hop reasoning and cross-graph information combination~\cite{Ji2022Survey}. Incorporating KGs with retrieval-augmented generation (RAG)~\cite{Lewis2020RAG}, question-answering systems can follow chains of relations rather than depending on text similarity in GraphRAG~\cite{Peng2024GraphRAGSurvey,Edge2024GraphRAG}. Recent methods~\cite{Hu2025ReMindRAG,Luo2025GFMRAG}  propose single traversal strategies; none compare multiple directions given hierarchical structures of knowledge. On the KG construction stage, LLMs perform zero-shot relation extraction~\cite{Wadhwa2023ACL} and speed up ontology creation~\cite{Shimizu2025AcceleratingKGOntology}, but the ontology could grow too large with overlapping entities and relations~\cite{Pan2023LLM}. Our main contributions are as follows:

\begin{itemize}
    \item A hierarchical KG construction pipeline with controlled ontology expansion is constructed using a two-layer ontology with 26 general, 14 history-specific types, an embedding centroid filter that reduces the number of prompts by $50\%$, and a five-step dual-LLM validator that prevents 87.5\% of duplicate relations. 

    \item An automatic benchmark generation pipeline from KGs, including a three-phase subgraph selection process such as SEED, ENRICH, BOUND generates 1,210 Vietnamese queries from 109~subgraphs with hierarchical relevance labels (i.e., 0/1/2), categorized by five query directions.

    \item Three graph traversal strategies, such as
    Top-Down, Horizontal, and Bottom-Up, are evaluated using a hierarchical structure along different directions.

    \item A multi-dimensional evaluation framework including 12 methods at 4 cutoff thresholds $K$, analyzes performance across 5 query directions, and the influences of the structural relations.
\end{itemize}

\section{Related Work}\label{sec:related}

\vspace{-0.2cm} 
\subsubsection{RAG and GraphRAG:}\label{sec:related-rag}

The RAG systems were designed with a generative retriever, and the LLM can produce its answers in retrieved documents~\cite{Lewis2020RAG}. Adding a knowledge graph to the pipeline, GraphRAG~\cite{Peng2024GraphRAGSurvey,Edge2024GraphRAG} enables the system to follow relational paths rather than rely on keyword overlap alone. Existing methods traverse the graph in different ways: ReMindRAG~\cite{Hu2025ReMindRAG} allows the LLM to decide which edges to follow, GFM-RAG~\cite{Luo2025GFMRAG} trains a graph foundation model for the task on hierarchical structures. RAPTOR~\cite{Sarthi2024RAPTOR} builds a retrieval tree through recursive abstractive summarization. Unlike existing works, our approach keeps the original document hierarchy and evaluates multiple traversal directions rather than a single bottom-up clustering strategy.

\vspace{-0.2cm} 
\subsubsection{Knowledge Graph Construction from Documents:}\label{sec:related-kgconstruct}

Recent works in ~\cite{Salman2024DocKG,Bian2025AutoMathKG,Lu2025KARMA} use LLMs to extract knowledge graphs from unstructured documents. Doc-KG~\cite{Salman2024DocKG} and AutoMathKG~\cite{Bian2025AutoMathKG} both combine embedding-based relation filtering with LLM extraction. LLMs can perform zero-shot relation extraction~\cite{Wadhwa2023ACL} and speed up ontology design~\cite{Shimizu2025AcceleratingKGOntology}, the expansion quickly produces duplicate types~\cite{Pan2023LLM}. The work in~\cite{Lu2025KARMA} uses multi-agent LLMs to enrich existing KGs, though enrichment differs from construction. Our proposed system will combine controlled expansion with dual-LLM validation to prevent the ontology from growing too large.
In the education domain, Ain et al.~\cite{Ain2025EduKGPipeline,Ain2025TopDownBottomUp} build a KG pipeline for educational content and compare top-down with bottom-up construction. Abu-Rasheed et al.~\cite{AbuRasheed2025KGCurriculum} uses LLMs to fill in missing relations in curriculum KGs. To the best of our knowledge, there is a lack of KGs built from Vietnamese textbooks.

\vspace{-0.2cm} 
\subsubsection{Benchmarks for KG-based Retrieval:}\label{sec:related-benchmark}

CRAG is the most popular RAG benchmark~\cite{Yang2024CRAG}, which has 4,409 question-answer pairs and evaluates flat text and entity-level question answering. Graded relevance labels, where a node can be partially relevant rather than simply relevant or not, are absent from existing benchmarks. 

Therefore, three open challenges remain and will be addressed in our paper: (i)~building a KG with controlled ontology expansion and dual-LLM validation for a low-resource language, (ii)~designing a retrieval benchmark with graded relevance labels and query direction categories, and (iii)~comparing traversal strategies on a hierarchical structure for Vietnamese history textbooks.

 \section{ Tree Knowledge Graph Construction
using Ontology Expansion} 


\begin{figure}[htb]
\centering
\includegraphics[width=\textwidth]{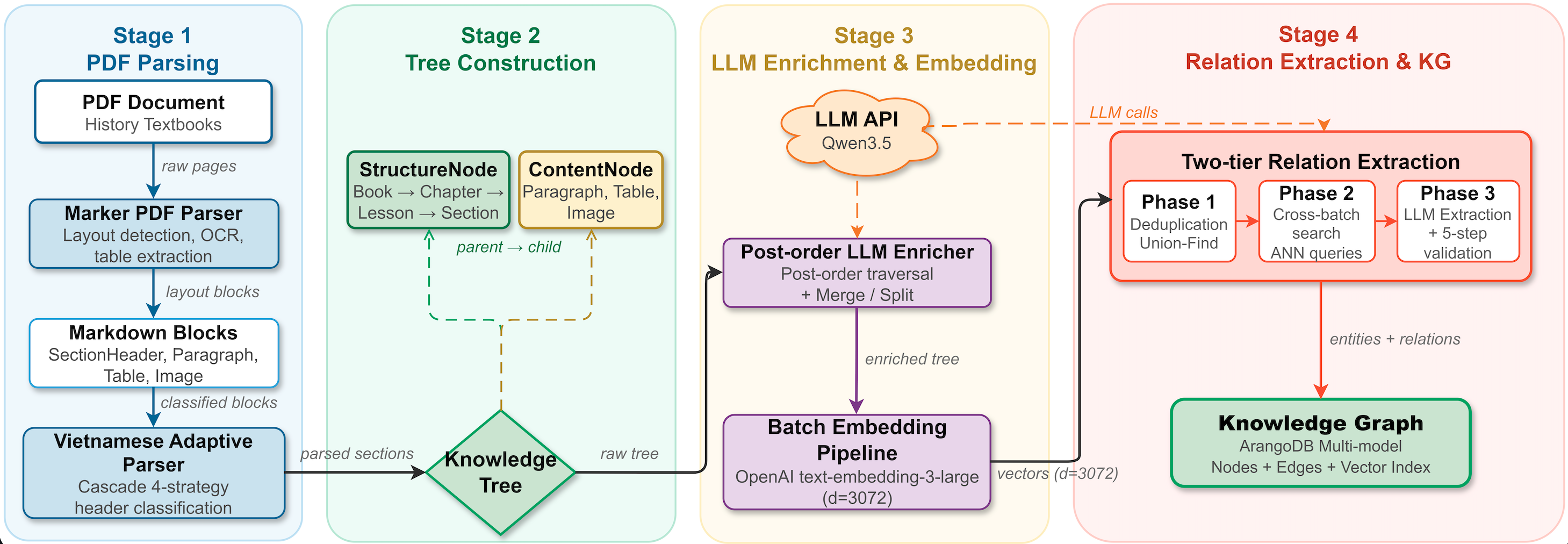}
\caption{Four-stage pipeline for hierarchical KG construction.}
\label{fig:pipeline}
\end{figure}

Figure~\ref{fig:pipeline} illustrates an overview of the proposed system, which has two parts: Part~(A) builds the KG through four stages, and Part~(B) evaluates retrieval using three graph traversal strategies. The four construction stages are:

\begin{itemize}
    \item \emph{Stage 1 - Document parsing}: We convert PDF files into an $n$-ary hierarchical tree using a deep learning-based layout analysis tool~\cite{Zhang2024DocParsingSurvey} and produce two kinds of nodes: \emph{structural nodes} (e.g., books, chapters, lessons, sections) and \emph{content nodes} (e.g., paragraphs, tables). The content nodes are initially split into small pieces; LLM-based merging of segments within the same structural node is therefore necessary to preserve context. In our design, the structural nodes will not be merged to keep the original structure of documents, but the content nodes could be merged to reduce duplication. 
    \item \emph{Stage 2 - Knowledge tree enrichment}: We perform post-order traversal to enhance each structural node with extra context based on the summary from its child content nodes.
    \item \emph{Stage 3 - Embedding generation}: Vector representations are created for nodes.
    \item \emph{Stage 4 - Relation extraction}: We developed a pipeline for relation extraction, including embedding filtering, LLM classification, and ontology.
\end{itemize}

The resulting KG is stored in ArangoDB~\cite{ArangoDB2024} (multi-model: document + graph), which serves both benchmark generation (Section~\ref{sec:benchmark}) and retrieval (Section~\ref{sec:strategies}).




\paragraph{Two-layer ontology:}
The ontology includes 40 initial relation types organized into two layers. The general layer contains 26 relation types that are reusable across all domains. The history-specific layer contains 14 relation types covering causal, temporal, historical figure, and other relations. The two-layer design allows the domain-specific layer to be replaced when changing to a different domain, while keeping both the general layer and the extraction pipeline. Each relation type is designed with a short description, keywords, node type constraints, and an embedding vector. The tier partitioning separates Tier~1 for structural nodes from Tier~2 for content nodes.

\vspace{-0.2cm}
\subsubsection{Stage 4 - Relation Extraction with three phases:}\label{sec:extraction}



\paragraph{Phase 1 - Pre-filtering:}
We first calculate the cosine similarity score $s$ for the node pair. The pairs having scores above the threshold $\tau_{\text{merge}}$ will be merged automatically or linked with an equivalence edge. Those falling between $\tau_{\text{rel}}$ and $\tau_{\text{merge}}$ go through the LLM for relation classification in Phase 2, and pairs below $\tau_{\text{rel}}$ stay separate. In this way, LLM focused on the middle range cosine similarity only.

\paragraph{Phase 2 - Ontology filtering:}
Given the node pair $(A, B)$, the embedding centroid is $\mathbf{c}_{AB} = \tfrac{1}{2}(\mathbf{e}_A + \mathbf{e}_B)$. The relevance score for relation type $R$ is:
\begin{equation}\label{eq:centroid}
    \text{score}(R) = {\mathbf{c}}_{AB} \cdot {\mathbf{e}}_R
\end{equation}
 where $\mathbf{e}$ are the embedding vectors of nodes or relations. To save the tokens and processing time, we employ the max-pooling strategy to identify the most relevant relation types of pairs $p$ in batch such as $\max_{p \in \text{batch}} \left({\mathbf{c}}_p \cdot {\mathbf{e}}_R\right)$.
In this way, our process can select the top 15 most relevant relation types from 40 types and reduce the prompt size by nearly 50\%.

\paragraph{Phase 3 - Dual LLM relation extraction:}
A weak LLM model receives batches of node pairs and 15 filtered relation candidates to extract relations. For each pair, the model either selects a type from the candidates, declares \textit{no relation}, or proposes a \textit{novel relation}. In our experiments, the weak LLM proposed many novel relation types per batch; however, nearly 95\% relations were almost similar to the existing 40 relations (e.g., \texttt{LED\_TO} vs.\ \texttt{CAUSED}). Without filtering, the ontology would introduce many new relations having almost similar meaning and cause the \emph{bloated ontology} problem. Hence, the novel relations need to be double-checked with the existing 40 relation types and filtered based on embedding similarity. 
Thereafter, a strong LLM model is used to validate the proposed novel relations.
First, the proposed relation of node pairs from the weak LLM will be either accepted or introduced a better relation, including the name, description, relation family, and constraints using the strong LLM model. Then the novel relations will be double-checked with the existing relations using the cosine similarity score with the threshold $\tau_{\text{onto}}$. For example, \texttt{CONTRIBUTED\_TO} and \texttt{PLAYED\_ROLE\_IN} that share the same meaning under different names.

\section{Graph Retrieval-Augmented Generation Strategies}\label{sec:strategies}
\begin{figure}[htb]
    \centering
    \includegraphics[width=\textwidth]{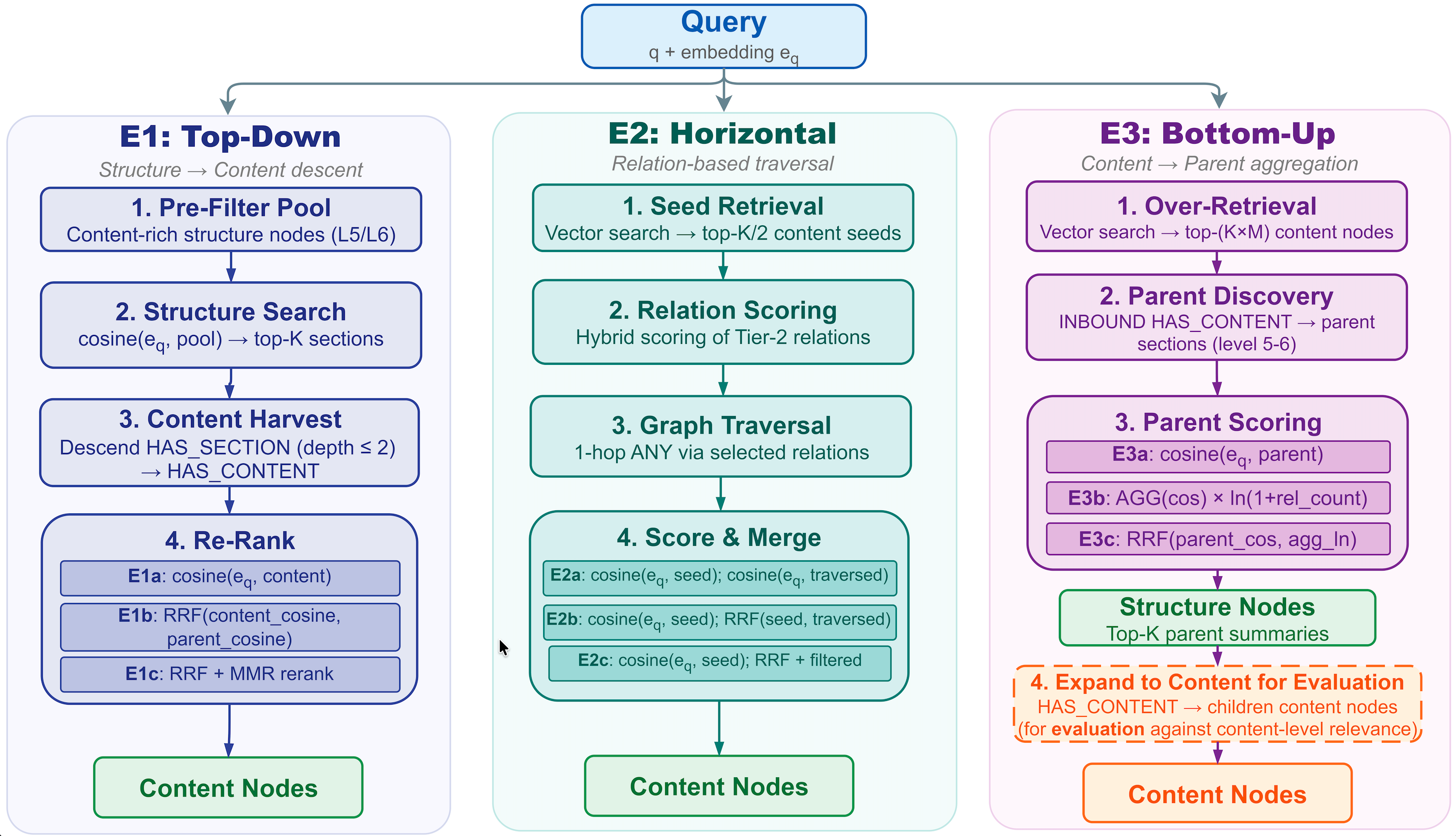}
    \caption{Three graph traversal strategies: Top-Down traverses from structural nodes down to content, Horizontal traverses via semantic relations, Bottom-Up aggregates from content up to parent nodes.}
    \label{fig:retrieval-strategies}
\end{figure}
\vspace{-0.2cm}
In this paper, we experimented with three Graph-RAG strategies for the constructed tree knowledge, as illustrated in Figure~\ref{fig:retrieval-strategies}. We also propose ranking scores based on the hierarchical position, parent-child relationships, and the semantic score between nodes.

Reciprocal Rank Fusion (RRF)~\cite{Cormack2009RRF} was applied to build the scores for strategies as follows:
\begin{equation}\label{eq:rrf}
    \text{RRF}(d) = \sum_{s \in S} \frac{1}{k + \text{rank}_s(d)}
\end{equation}
where $k$ is a smoothing constant, used to merge multiple ranking criteria.

\vspace{-0.2cm} 
\subsubsection{Top-Down Strategy:}\label{sec:e1}
The strategy first identifies the best-matching sections using a semantic summary of each structural node based on embedding similarity. Then it traverses downward to collect all content nodes in the subtree of sections.
Three ranking scores are evaluated: \textbf{E1a} uses a cosine similarity score, \textbf{E1b} uses the RRF score given \texttt{seed\_score}, and \textbf{E1c} extends E1b with MMR~\cite{Carbonell1998MMR} to enhance the diversity.

\vspace{-0.2cm} 
\subsubsection{Horizontal Strategy:} \label{sec:e2}

Many queries require connecting scattered information across different sections. For example, a cause might appear in one section while its consequence events appear in another. The Horizontal strategy addresses the problem by traversing semantic relation edges between branches of the tree. The process starts with a vector search to select seed nodes and the corresponding \texttt{seed\_score} based on the cosine similarity score. Then, it filters to select the relevant relation types and traverses one hop to reach cross-branch nodes (i.e., extended nodes). Among all of the selected nodes, the strategy performs the following three rankings to select the top-K relevant nodes.
First, \textbf{E2a} uses only cosine similarity between the content nodes, including the seed and extended nodes. Second,
\textbf{E2b} uses the RRF score given \texttt{seed\_score}, the cosine similarity of the extended content nodes. Third, \textbf{E2c} is a restricted variant when keeping only the top-3 relation types.

\vspace{-0.2cm} 
\subsubsection{Bottom-Up Strategy:}\label{sec:e3}

Given the query, when the content nodes have a high similarity score, the shared parent structural node is likely relevant. The approach is especially useful when the parent summary is too general to match the query directly.
The main factor for ranking is the threshold-based child aggregation. We denote $\mathcal{C}(p)$ as the set of child nodes belonging to the parent node $p$, and $s_c = \text{sim}(\mathbf{q}, \mathbf{e}_c)$ is the cosine similarity between the query and child node $c$. The threshold $\tau_p$ is computed as the median of all child scores belonging to the parent node $p$. Only children whose score is greater than or equal to $\tau_p$ are selected.
The process starts with a vector search to retrieve candidate content nodes, then follows inbound edges to identify parent nodes using the following three variants. First, \textbf{E3a} uses cosine similarity based on the semantic summary to select top-K parent nodes. Second, \textbf{E3b} applies the aggregated score of the parent node $p$ based child scores:

\begin{equation}\label{eq:avglog}
    \text{agg\_score}(p)
    = \frac{\sum_{c \in C(p)} s_c}{|\mathcal{C}(p)|}
    \;\times\;
    \ln\!\Bigl(1 + \bigl|\{c \in C(p) : s_c \geq \tau_p\}\bigr|\Bigr)
\end{equation}
The first term captures the average cosine quality of all children under parent $p$. The second term introduces the logarithmic scaling factor based on the number of relevant children, thereby avoiding the explosion of dense parents.
Third, \textbf{E3c} utilizes RRF using the parent cosine and \text{agg\_score}(p). 

\section{Benchmark Generation Pipeline}\label{sec:benchmark}

\begin{figure}[htb]
    \centering
    \includegraphics[width=\textwidth]{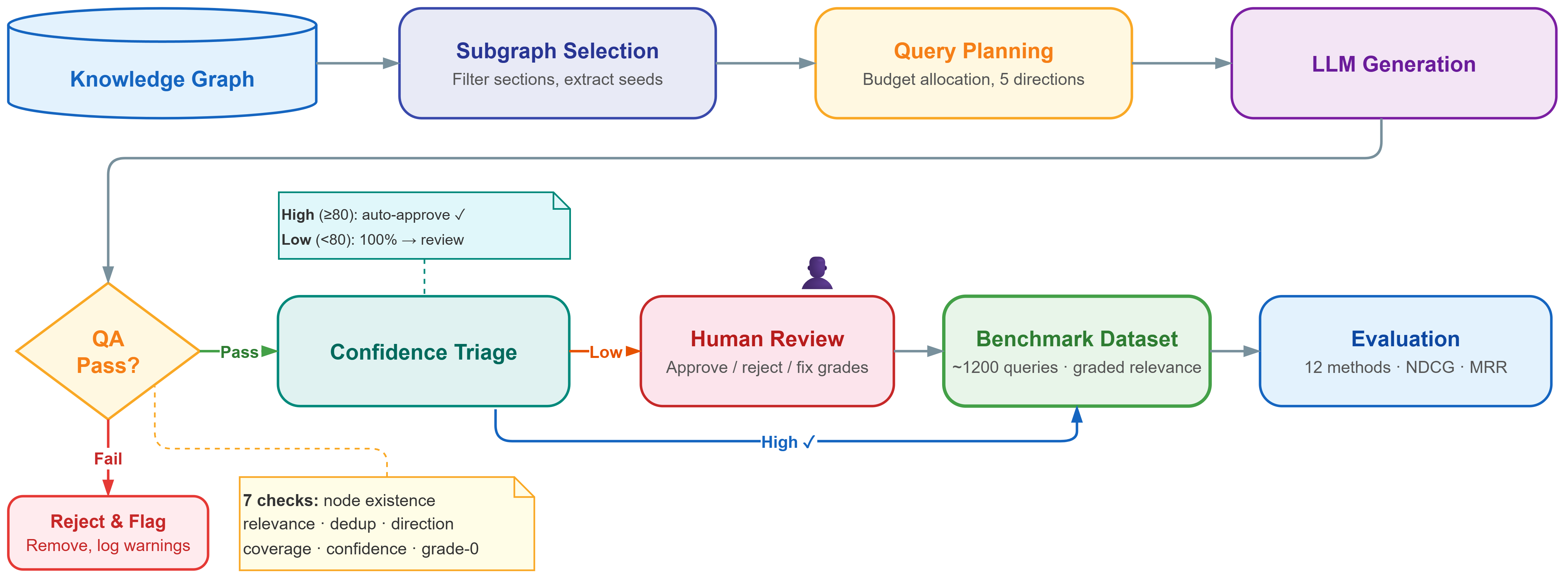}
    \caption{Benchmark generation pipeline: SEED--ENRICH--BOUND subgraph selection, LLM-based query generation along 5 directions, 7-step quality control $\rightarrow$ 1,210 Vietnamese queries.}
    \label{fig:benchmark-pipeline}
\end{figure}

To validate three Graph-RAG strategies, we construct the benchmark using the proposed pipeline in Figure~\ref{fig:benchmark-pipeline}. In this way, we can convert the constructed KG into a benchmark dataset with corresponding graded relevance labels for five query directions.

\vspace{-0.5cm}

\vspace{-0.2cm} 
\subsubsection{Three-Phase Subgraph Selection:}

From the KG stored in ArangoDB, the system selects all of the L5-level structural nodes that have at least one child content node. Each L5 candidate is assembled through three phases:
\begin{itemize}
\item SEED: Traversal from the selected L5 node to collect core content nodes.

\item ENRICH: From each core content node, one-hop traversal via Tier-2 semantic relations discovers content nodes belonging to other sections and selects at most two sections ranked by the number of relations between content nodes. Then we retrieve the parent nodes and child nodes that belong to them. 

\item BOUND: We filter to select the relevant relations among collected nodes (i.e., core and enriched nodes), forming a closed subgraph. 
\end{itemize}

\vspace{-0.5cm} 
\subsubsection{Five Query Directions:}

The benchmark categorizes queries along five directions reflecting distinct information needs.
Neutral ({NT}) queries perform a single-fact lookup, require only one node serving as a baseline.
Top-down ({TD}) queries synthesize multiple nodes within the same section.
Horizontal ({HZ}) queries relate to the information from different sections.
Bottom-up ({BU}) queries target summaries or themes at the section level.
Finally, Cross-section ({CS}) queries compare content from two distinct sections.

\vspace{-0.2cm} 
\subsubsection{Query Generation and Quality Control:}

The LLM generates JSON including the query, a list of relevant nodes and their relevant level 0/1/2, direction, and a confidence score. A seven-step QA verification for each output will help to check the nodes existence, content relevance exceeding a cosine threshold, remove the duplicates, remove non-relevant nodes, and validate directions given their quotas. High-confidence outputs receive a 10\% spot-check; medium-confidence outputs are reviewed at a 50\% rate; every low-confidence item is reviewed in full.

\section{Experiments and Results}\label{sec:experiment}


\vspace{-0.2cm} 
\subsection{Benchmark dataset}

The KG is constructed from three Vietnamese high school History textbooks of the Canh Dieu edition for Grades 10, 11, and 12, totaling approximately 400 pages and covering world and Vietnamese history from ancient times to the modern era, following the 2018 General Education Curriculum.
In this work, 3 LLM models were employed across the pipelines. The weak LLM model Qwen3.5 397B A17B~\cite{Qwen3TechReport2025} handles bulk relation extraction, accounting for over 95\% of the workload, and also synthesizes summaries for structural nodes; the strong LLM model Claude 4.6 Sonnet~\cite{Anthropic2026Sonnet46} validates novel relation types. The embedding model text-embedding-3-large~\cite{OpenAI2024Embedding3} with $d{=}3072$ dimensions generates vector representations for all nodes and ontology entries.

The automated generation pipeline described in Section~\ref{sec:benchmark} produces 1,210 Vietnamese queries from 109 subgraphs. After quality verification, 1,182 queries are clean and 28 are flagged. The relevance label distribution is as follows: grade-0, meaning not relevant, accounts for 2,400 labels at 36.9\%; grade-1, meaning partially relevant, accounts for 1,143 labels at 17.6\%; and grade-2, meaning directly relevant, accounts for 2,968 labels at 45.6\%. Query direction distribution: TD with 321 queries at 26.5\%, HZ with 363 at 30.0\%, BU with 290 at 24.0\%, NT with 136 at 11.2\%, and CS with 100 at 8.3\%.

\textbf{Metrics:}
We evaluate all strategies at cutoffs $K = \{1, 3, 5, 10\}$. We use NDCG@K~\cite{Jarvelin2002NDCG} as the main metric because it works well with our three-level graded relevance labels (i.e., grade 0/1/2). Recall@K and Precision@K measure coverage and exactness. MRR measures how quickly the first correct result appears.

\textbf{Baselines:}
We evaluated three baselines in this paper: \textbf{B1} performs vector search with cosine similarity on ArangoSearch, \textbf{B2} uses BM25~\cite{Robertson2009BM25} for lexical search, \textbf{B3} combines both methods through RRF.

\vspace{-0.2cm} 
\subsubsection{Graph Construction Results:}\label{sec:res-kg}

Table~\ref{tab:tree} gives an overview of the graph. It contains 750 nodes, including 3 root, 263 structural, and 484 content nodes, connected by 5,372 edges of which 1,031 are tree-structural and 4,341 are semantic. The four structural levels span 15 chapters, 28 lessons, 109 sections, and 111 subsections.

\begin{table}[tb]
\caption{Knowledge graph statistics and distribution of the 10 most frequent semantic relation types}
\label{tab:tree}
\centering\small
\begin{tabular}{lr@{\hskip 2em}lr}
\toprule
\textbf{Graph component} & \textbf{Counts} & \textbf{Top 10 Relation } & \textbf{\#Edges} \\
\midrule
Root nodes (books) & 3 & \textsc{Contextualizes} & 830 \\
Structural nodes & 263 & \textsc{Refers\_To} & 620 \\
Content nodes & 484 & \textsc{Contrasts} & 444 \\
\textbf{Total nodes} & \textbf{750} & \textsc{Exemplifies} & 345 \\
\cmidrule{1-2}
Structural edges & 1,031 & \textsc{Elaborates} & 324 \\
Semantic edges & 4,341 & \textsc{Preceded\_By} & 319 \\
\textbf{Total edges} & \textbf{5,372} & \textsc{Related\_To} & 253 \\
 &  & \textsc{Supports} & 230 \\
 &  & \textsc{Part\_Of} & 128 \\
 & & \textsc{Followed\_By} & 115 \\
\bottomrule
\end{tabular}
\end{table}



\vspace{-0.2cm} 
\subsection{Overall Retrieval Ranking}\label{sec:res-ranking}

In the experiments, we evaluate and rank 12 methods, including 9 Graph-RAG variant strategies and 3 baselines, using NDCG@10 across all 1,210 queries as shown in Table~\ref{tab:overall-ranking}. 

\begin{table}[htb]
\centering
\caption{Overall ranking of 12 methods at K=10 on 1,210 queries. }
\label{tab:overall-ranking}
\small
\begin{tabular}{rllcccc}
\toprule
\# & ID & Name & NDCG & Recall & Precision & MRR \\
\midrule
1 & \textbf{E1b} & TopDown RRF & \textbf{0.8397} & \textbf{0.9053} & \textbf{0.3040} & 0.8967 \\
2 & \textbf{E3c} & BottomUp RRF & 0.8122 & 0.8307 & 0.2711 & \textbf{0.9331} \\
3 & \textbf{E3b} & BottomUp AVG\_LOG & 0.8110 & 0.8300 & 0.2704 & 0.9310 \\
4 & \textbf{E1a} & TopDown Cosine & 0.8075 & 0.8284 & 0.2721 & 0.9288 \\
5 & \textbf{E1c} & TopDown MMR & 0.8065 & 0.8268 & 0.2704 & 0.9299 \\
6 & \textbf{E3a} & BottomUp Parent & 0.8036 & 0.8236 & 0.2710 & 0.9253 \\
7 & \textbf{E2a} & Horiz. Cosine & 0.8009 & 0.8090 & 0.2634 & 0.9298 \\
8 & B1 & Vector-Only & 0.7923 & 0.8179 & 0.2637 & 0.9301 \\
9 & B3 & Hybrid RRF & 0.7806 & 0.7903 & 0.2523 & 0.9262 \\
10 & \textbf{E2c} & Horiz. Filtered & 0.7539 & 0.7369 & 0.2812 & 0.9256 \\
11 & \textbf{E2b} & Horiz. RRF & 0.7343 & 0.7422 & 0.2382 & 0.9190 \\
12 & B2 & BM25 & 0.6770 & 0.6791 & 0.2112 & 0.8508 \\
\bottomrule
\end{tabular}
\end{table}

E1b — The TopDown RRF strategy outperforms the other retrieval methods for all metrics  due to the prevalence of top-down query styles of the benchmark datasets. In practice, many queries are inherently inclusive from structured textbook sections and align naturally with the hierarchical approach. Consequently, TopDown RRF achieves an NDCG@10 of 0.8397 and a Recall@10 of 0.9053, representing improvements of 4.7 and 8.7 percentage points, respectively, over the B1 baseline. Most of Graph-RAG variants outperform the three baselines, demonstrating the usefulness of structural information in the RAG process.
Besides, Bottom-Up E3c achieves the highest MRR of any method at 0.9331, meaning it reliably places the best result first, but it ranks second on NDCG. The score aggregation mechanism from child nodes described in Equation~\ref{eq:avglog} is effective to identify the single most relevant section.
E2a - Horizontal strategy slightly outperforms B1 by 0.86 percentage points in NDCG through reaching cross-branch nodes, but E2b drops 6.7 percentage points relative to E2a. There are some overlapping information between Seed node and extended nodes introducing noise in RRF.


\vspace{-0.2cm} 
\subsubsection{Analysis for different K:}\label{sec:res-multi-k}

Table~\ref{tab:multi-k} compares NDCG at four cutoff values for the best method in each strategy alongside two baselines.
\begin{table}[htb]
\centering
\caption{NDCG, Recall, and Precision at $K{=}1, 3, 5, 10$ for representative methods.}
\label{tab:multi-k}
\resizebox{\columnwidth}{!}{%
\begin{tabular}{l ccc ccc ccc ccc}
\toprule
& \multicolumn{3}{c}{$K{=}1$}
& \multicolumn{3}{c}{$K{=}3$}
& \multicolumn{3}{c}{$K{=}5$}
& \multicolumn{3}{c}{$K{=}10$} \\
\cmidrule(lr){2-4}\cmidrule(lr){5-7}\cmidrule(lr){8-10}\cmidrule(lr){11-13}
Method
& NDCG & Rec. & Pre.
& NDCG & Rec. & Pre.
& NDCG & Rec. & Pre.
& NDCG & Rec. & Pre. \\
\midrule
\textbf{E1b} (TD RRF)
& 0.820 & 0.308 & 0.836
& \textbf{0.758} & \textbf{0.627} & \textbf{0.640}
& \textbf{0.786} & \textbf{0.774} & \textbf{0.497}
& \textbf{0.840} & \textbf{0.905} & \textbf{0.304} \\
\textbf{E3c} (BU RRF)
& \textbf{0.881} & 0.345 & \textbf{0.893}
& 0.754 & 0.596 & 0.596
& 0.759 & 0.699 & 0.439
& 0.812 & 0.831 & 0.271 \\
\textbf{E2a} (HZ Cos)
& 0.871 & 0.344 & 0.883
& 0.745 & 0.590 & 0.584
& 0.756 & 0.698 & 0.434
& 0.801 & 0.809 & 0.263 \\
B1 (Vector)
& 0.871 & 0.344 & 0.883
& 0.724 & 0.590 & 0.584
& 0.746 & 0.698 & 0.434
& 0.792 & 0.818 & 0.264 \\
B3 (Hybrid)
& 0.867 & \textbf{0.347} & 0.881
& 0.718 & 0.563 & 0.548
& 0.726 & 0.660 & 0.405
& 0.781 & 0.790 & 0.252 \\
\bottomrule
\end{tabular}%
}
\end{table}
Bottom-Up strategy with $K=1$ results in the highest score because the child-node can produce the single best section. From K$\geq$3, Top-Down obtains the highest score because the seed\_score propagated downward helps all children in the right section and obtains sufficient coverage in many queries from the benchmark. 
The gap between graph strategies and baselines grows as K increases. At $K=1$, B1 matches E2a at 0.871; at $K=10$, E1b has a higher score than B1 by 4.7~point. Note that the higher $K$ strategies produce a lower NDCG@10 and Precision score compared to $K=1$; however, the coverage has been extended as the increment of the recall scores. Structural information becomes more influenced when $K$ increases.

\vspace{-0.2cm} 
\subsection{Strategy $\times$ Query Direction Matrix}\label{sec:res-direction}

Table~\ref{tab:direction-matrix} presents NDCG@10 broken down by five query directions. The direction-level breakdown shows large differences between strategies:

\begin{table}[htb]
\centering
\caption{NDCG@10 by query direction, $K=10$.}
\label{tab:direction-matrix}
\small
\begin{tabular}{lccccc}
\toprule
Method & TD & HZ & BU & NT & CS \\
\midrule
B1 (Vector) & 0.762 & 0.832 & 0.716 & 0.925 & 0.755 \\
B3 (Hybrid) & 0.755 & 0.826 & 0.680 & \textbf{0.962} & 0.742 \\
\textbf{E1b} (TD RRF) & \textbf{0.876} & 0.826 & \textbf{0.878} & 0.741 & \textbf{0.799} \\
\textbf{E2a} (HZ Cos) & 0.771 & \textbf{0.861} & 0.709 & 0.923 & 0.777 \\
\textbf{E3c} (BU RRF) & 0.791 & 0.858 & 0.744 & 0.897 & 0.797 \\
\bottomrule
\end{tabular}
\end{table}

\begin{itemize}
    \item \emph{TD:} E1b reaches 0.876, which is 11.4~point above B1.
    \item \emph{HZ:} E2a leads at 0.861 by finding causally or temporally related nodes in other sections.
    \item \emph{BU:} E1b scores 0.878 and surpasses even E3c at 0.744.
    \item \emph{NT:} B3 reaches 0.962, the highest score in the entire table, while E1b drops to 0.741 because structural traversal spreads results across irrelevant nodes.
    \item \emph{CS:} The most challenged direction for all methods, as no strategy exceeds 0.80.
\end{itemize}




\section{Discussions}\label{sec:conclusion}

In this paper, we proposed a hierarchical KG construction pipeline with controlled ontology expansion and compared three Graph-RAG strategies from Vietnamese history textbooks. The preliminary results exhibit a promising direction for future Graph-RAG assisted educational agents to produce confident and high-quality answers. 
While Graph-RAG strategies utilizing structural information could produce better and inclusive answers for complex queries that span multiple sections, certain limitations remain. Due to limited capacity, our study was validated on high school history textbooks using an automated benchmark construction process, resulted in a relatively small scale of nodes and relations. Expert-annotated benchmarks for both extraction and retrieval evaluation are also important to enhance the consistency of experimental studies. The use of hand-tuned thresholds may affect the variant performance of graph retrieval systems. In future work, we will extend the pipeline to other domains and experiment more efficient retrieval strategies capable of handling increasingly complex queries.

\textbf{Acknowledgment}
This research is funded by Vietnam - Korea University of Information and Communication Technology under project number ĐHVH-2026-23.

\bibliographystyle{splncs04}
\bibliography{references}

@inproceedings{Lewis2020RAG,
  author    = {Patrick Lewis and Ethan Perez and others},
  title     = {Retrieval-Augmented Generation for Knowledge-Intensive {NLP} Tasks},
  booktitle = {Advances in Neural Information Processing Systems (NeurIPS)},
  volume    = {33},
  pages     = {9459--9474},
  year      = {2020}
}

@article{Peng2024GraphRAGSurvey,
  author    = {Boci Peng and Yun Zhu and others},
  title     = {Graph Retrieval-Augmented Generation: A Survey},
  journal   = {ACM Transactions on Information Systems},
  volume    = {44},
  number    = {2},
  pages     = {1--52},
  year      = {2025},
}

@article{Edge2024GraphRAG,
  author    = {Darren Edge and Ha Trinh and others},
  title     = {From Local to Global: {A} Graph {RAG} Approach to Query-Focused Summarization},
  journal   = {arXiv preprint arXiv:2404.16130},
  year      = {2024}
}

@inproceedings{Hu2025ReMindRAG,
  author    = {Yikuan Hu and Jifeng Zhu and others},
  title     = {{ReMindRAG}: Low-Cost {LLM}-Guided Knowledge Graph Traversal for Efficient {RAG}},
  booktitle = {Advances in Neural Information Processing Systems (NeurIPS)},
  year      = {2025}
}

@inproceedings{Luo2025GFMRAG,
  author    = {Linhao Luo and Zicheng Zhao and others},
  title     = {{GFM-RAG}: Graph Foundation Model for Retrieval Augmented Generation},
  booktitle = {Advances in Neural Information Processing Systems (NeurIPS)},
  year      = {2025}
}

@inproceedings{Sarthi2024RAPTOR,
  author    = {Parth Sarthi and Salman Abdullah and others},
  title     = {{RAPTOR}: Recursive Abstractive Processing for Tree-Organized Retrieval},
  booktitle = {The Twelfth International Conference on Learning Representations ({ICLR})},
  year      = {2024}
}

@inproceedings{Yang2024CRAG,
  author    = {Xiao Yang and Kai Sun and others},
  title     = {{CRAG} -- Comprehensive {RAG} Benchmark},
  booktitle = {Advances in Neural Information Processing Systems (NeurIPS), Datasets and Benchmarks Track},
  year      = {2024}
}

@article{Ji2022Survey,
  author    = {Shaoxiong Ji and Shirui Pan and others},
  title     = {A Survey on Knowledge Graphs: Representation, Acquisition, and Applications},
  journal   = {IEEE Transactions on Neural Networks and Learning Systems},
  volume    = {33},
  number    = {2},
  pages     = {494--514},
  year      = {2022}
}

@article{Salman2024DocKG,
  author    = {Muhammad Salman and Armin Haller and others},
  title     = {Doc-{KG}: Unstructured Documents to Knowledge Graph Construction, Identification and Validation with Wikidata},
  journal   = {Expert Systems},
  volume    = {41},
  number    = {9},
  year      = {2024}
}

@article{Bian2025AutoMathKG,
  author    = {Rong Bian and Yu Geng and others},
  title     = {{AutoMathKG}: The Automated Mathematical Knowledge Graph Based on {LLM} and Vector Database},
  journal   = {Computational Intelligence},
  volume    = {41},
  number    = {4},
  year      = {2025},
}

@inproceedings{Ain2025EduKGPipeline,
  author    = {Qurat Ul Ain and Mohamed Amine Chatti and others},
  title     = {An Optimized Pipeline for Automatic Educational Knowledge Graph Construction},
  booktitle = {Proceedings of the International Joint Conference on Knowledge Graphs (IJCKG)},
  year      = {2025},
}

@inproceedings{Ain2025TopDownBottomUp,
  author    = {Qurat Ul Ain and Mohamed Amine Chatti and others},
  title     = {Top-Down vs.\ Bottom-Up Approaches for Automatic Educational Knowledge Graph Construction in {CourseMapper}},
  booktitle = {Proceedings of the European MOOC Stakeholder Summit (EMOOCs)},
  year      = {2025},
  pages     = {119--129}
}

@inproceedings{AbuRasheed2025KGCurriculum,
  author    = {Hasan Abu-Rasheed and Constance Jumbo and others},
  title     = {{LLM}-Assisted Knowledge Graph Completion for Curriculum and Domain Modelling in Personalized Higher Education Recommendations},
  booktitle = {Proceedings of the IEEE Global Engineering Education Conference (EDUCON)},
  pages     = {1--5},
  year      = {2025}
}

@article{Zhang2024DocParsingSurvey,
  author    = {Zhang, Qintong and Wang, Bin and Huang, Victor Shea-Jay and others},
  title     = {Document Parsing Unveiled: Techniques, Challenges, and Prospects for Structured Information Extraction},
  journal   = {arXiv preprint arXiv:2410.21169},
  year      = {2024}
}

@inproceedings{Wadhwa2023ACL,
  author    = {Somin Wadhwa, Silvio Amir, Byron C. Wallace},
  title     = {Revisiting Relation Extraction in the era of Large Language Models},
  booktitle = {Proceedings of the 61st Annual Meeting of the ACL},
  pages     = {15566--15589},
  year      = {2023}
}

@article{Pan2023LLM,
  author    = { Jeff Z. Pan ORCID-Logo, Simon Razniewski and others},
  title     = {Large Language Models and Knowledge Graphs: Opportunities and Challenges},
  journal   = {Transactions on Graph Data and Knowledge (TGDK)},
  volume    = {1},
  number    = {1},
  pages     = {2:1--2:38},
  year      = {2023}
}

@article{Shimizu2025AcceleratingKGOntology,
  author    = {Shimizu, Cogan and Hitzler, Pascal},
  title     = {Accelerating Knowledge Graph and Ontology Engineering with Large Language Models},
  journal   = {Journal of Web Semantics},
  volume    = {85},
  pages     = {100862},
  year      = {2025}
}

@inproceedings{Lu2025KARMA,
  author    = {Lu, Yuxing and Wu, Wei and Zhao, Xukai and
               Peng, Rui and Wang, Jinzhuo},
  title     = {{KARMA}: Leveraging Multi-Agent {LLMs} for Automated
               Knowledge Graph Enrichment},
  booktitle = {Advances in Neural Information Processing Systems (NeurIPS)},
  year      = {2025}
}

@inproceedings{Cormack2009RRF,
  author    = {Gordon V. Cormack, Charles L A Clarke, Stefan Buettcher},
  title     = {Reciprocal Rank Fusion Outperforms Condorcet and Individual Rank Learning Methods},
  booktitle = {Proceedings of the 32nd International ACM SIGIR Conference on Research and Development in Information Retrieval},
  pages     = {758--759},
  year      = {2009}
}

@article{Robertson2009BM25,
  author    = {Robertson, Stephen and Zaragoza, Hugo},
  title     = {The Probabilistic Relevance Framework: {BM25} and Beyond},
  journal   = {Foundations and Trends in Information Retrieval},
  volume    = {3},
  number    = {4},
  pages     = {333--389},
  year      = {2009}
}

@inproceedings{Carbonell1998MMR,
  author    = {Carbonell, Jaime and Goldstein, Jade},
  title     = {The Use of {MMR}, Diversity-Based Reranking for Reordering Documents and Producing Summaries},
  booktitle = {Proceedings of the 21st Annual International ACM SIGIR Conference},
  pages     = {335--336},
  year      = {1998}
}

@misc{ArangoDB2024,
  author    = {{ArangoDB GmbH}},
  title     = {{ArangoDB}: Multi-Model Database for Graph, Document, Key-Value, and Search},
  year      = {2024}
}

@inproceedings{Nguyen2020PhoBERT,
  author    = {Nguyen, Dat Quoc and Nguyen, Anh Tuan},
  title     = {{PhoBERT}: Pre-trained Language Models for {Vietnamese}},
  booktitle = {Findings of the Association for Computational Linguistics: EMNLP 2020},
  pages     = {1037--1042},
  year      = {2020}
}

@article{Jarvelin2002NDCG,
  author    = {J{\"a}rvelin, Kalervo and Kek{\"a}l{\"a}inen, Jaana},
  title     = {Cumulated Gain-Based Evaluation of {IR} Techniques},
  journal   = {ACM Transactions on Information Systems},
  volume    = {20},
  number    = {4},
  pages     = {422--446},
  year      = {2002}
}

@misc{Qwen3TechReport2025,
  title         = {Qwen3 Technical Report},
  author        = {Chujie Zheng and Yuzhang Jiang and Zihan Xu and others},
  year          = {2025},
  eprint        = {2505.09388},
  archivePrefix = {arXiv},
  primaryClass  = {cs.CL}
}

@techreport{Anthropic2026Sonnet46,
  title        = {System Card: {Claude Sonnet} 4.6},
  author       = {{Anthropic}},
  institution  = {Anthropic PBC},
  year         = {2026},
  month        = {2},
  url          = {https://www.anthropic.com/system-cards}
}

@misc{OpenAI2024Embedding3,
  title        = {text-embedding-3-large},
  author       = {{OpenAI}},
  year         = {2024},
  howpublished = {\url{https://platform.openai.com/docs/models/text-embedding-3-large}},
}

\end{document}